# Clinical Relation Extraction Using Transformer-based Models

Authors: Xi Yang, PhD[1,2]
Zehao Yu, MS[1]
Yi Guo, PhD[1,2]
Jiang Bian, PhD[1,2]
Yonghui Wu, PhD[1,2]

Affiliation of the authors: [1]Department of Health Outcomes and Biomedical Informatics, College of Medicine, University of Florida, Gainesville, Florida, USA

[2]Cancer Informatics and eHealth core, University of Florida Health Cancer Center, Gainesville, Florida, USA

Corresponding author: Yonghui Wu, PhD
Clinical and Translational Research Building
2004 Mowry Road, PO Box 100177
Gainesville, FL, USA, 32610
Phone: 352-294-8436
Email: yonghui.wu@ufl.edu

Word count:

# ABSTRACT

## Introduction

Clinical relation extraction (RE) is a fundamental natural language processing (NLP) task to identify relations between clinical concepts from clinical narratives, which can subsequently help construct comprehensive patient profiles. Recently, transformer-based models have achieved state-of-the-art performances in many NLP benchmark tasks in the general domain; and there is an increasing interest in applying transformers for clinical RE. However, most existing transformer-based clinical RE studies have been heavily focused on the BERT architecture; and there is no study that has systematically examined other emerging transformer architectures such as RoBERTa and XLNet. Further, many key techniques of using transformers for clinical RE, such as how to effectively use the representations generated by transformers and how to handle cross-sentence relations remain to be challenging.

## Methods

This study explored 3 mainstream transformer architectures, including BERT, RoBERTa, and XLNet, for clinical RE. We systematically examined critical aspects of using transformers for clinical RE, including classification strategies (i.e., binary classification vs. multi-class classification), methods to handle cross-sentence relations, and strategies to effectively combine the representations generated by transformers for relation classification. We compared the 3 transformer architectures using 2 publicly available clinical RE datasets from 2018 MADE1.0 and 2018 n2c2 challenges.

**Results and Conclusion**

Two transformers achieved state-of-the-art performances, including RoBERTa, which achieved the best performance of 0.8959 (compared to the best F1-score of 0.8684 in challenge) for the 2018 MADE1.0 dataset, and XLNet, which achieved 0.9610 for the 2018 n2c2 dataset. When using transformer-based models for RE, the binary classification strategy is better than the multi-class strategy. Among the 3 transformers, RoBERTa and XLNet models overperform BERT for clinical RE. Our results also suggest that both the representations of the classification tokens and all entity markers generated by transformer models can help clinical RE. This study demonstrated the efficiency of transformer-based models for clinical RE. We believe this work will improve current practice in using transformers for clinical RE and other related NLP studies in the biomedical domain. Our methods and pretrained transformers are publicly available at https://github.com/uf-hobi-informatics-lab/ClinicalTransformerRelationExtraction.

# INTRODUCTION

Electronic Health Record (EHR) data has been widely used for clinical and translational research as it contains rich, longitudinal patient information [1–3]. As the emerging of artificial intelligence (AI) and data science, there has been an increasing interest to apply statistical and machine learning models to assess or predict health outcomes using EHRs. A critical step of using EHRs is to generate a complete picture of the patient's medical history with rich data elements from not only structured EHRs [4] (e.g., diagnosis codes) but also clinical narratives (e.g., physician notes and discharge summaries). Clinical narratives contain rich information that often does not exist in structured EHR data, such as family history, disease severity, adverse drug events, and social, behavioral determinants of health [5–7]. Natural language processing (NLP) has been applied to extract various information from clinical narratives to generate a comprehensive patient profile that is accessible to downstream AI and statistical models. Currently, NLP studies in the medical domain is heavily focused on information extraction (IE), including clinical concepts and relations between the clinical concepts [8]. For example, the clinical NLP community has organized a number of open challenges with shared tasks on both concept extraction and relation extraction (RE) [9–12]. And, there are numerous studies applying various NLP methods including traditional machine learning and the emerging deep learning models for clinical concept extraction (i.e., named entity recognition [NER]) in the medical domain [13,14]. The performance of clinical concept extraction models have greatly improved in the recent decade. For example, one of our clinical concept extraction models based on the transformer architectures outperformed the wining system using traditional conditional random fields (CRFs) in the 2010 i2b2 challenge by ~5% [15]. However, learning to extract relation is still challenging as it is a non-trivial task that has to consider the combinations among all clinical concepts.

RE is a fundamental NLP task that aims to establish semantic connections among concepts in a document [16]. Both rule-based and machine learning-based approaches for RE have been widely examined in the past [17]. Recently, deep learning models such as Recurrent Neural Networks (RNN), have demonstrated superior performances compared to the rule-based and traditional machine learning-based models [18,19]. More recently, the transformer architecture, built solely with a self-attention mechanism, has been introduced for language modeling and greatly advanced almost every NLP subfield [20]. Numerous transformer-based language models with different pretraining strategies have been developed and continue pushing the state-of-the-art performances on various NLP benchmark tests in the general domain [21–24]. Although there are a few studies explored transformer-based models for biomedical and clinical information extraction [15,25–31], there is no study that has systematically examined the mainstream transformer architectures for RE in the medical domain.

This is the first study that systematically explored transformer-based models for RE in the medical domain and developed an open-source NLP package to facilitate the use of transformers for RE from clinical narratives. We systematically examined 3 mainstream transformer-based architectures, including the bidirectional encoder representations from transformers (BERT) [32], RoBERTa [33], and XLNet [34] using two relation extraction benchmarks developed by the 2018 MADE1.0 [10] and 2018 n2c2 [11] challenges. For each transformer, we examined three different settings including two variants based on model's parameter sizes: BASE and LARGE, and the clinical model that is pretrained with clinical corpus. We compared all models using the standard precision, recall and F1-score at both individual category level (see supplement material) and micro-averaged level. The clinical RE package with 3 pretrained clinical transformer models is released as an open-source package (https://github.com/uf-hobi-informatics-lab/ClinicalTransformerRelationExtraction) to facilitate clinical RE using transformers.

## BACKGROUND

Information extraction (IE) is an NLP task that aims to automatically identify and encode predefined facts, events, and their relationships from text [35,36]. In the medical domain, IE is a key component of many downstream clinical applications, such as clinical decision support, computable phenotyping, and cohort discovery [8]. Clinical IE consists of two fundamental tasks: (1) clinical concepts extraction - identifying medical concepts from clinical text, and (2) clinical RE - establish semantic connections among the concepts. RE is the key step towards constructing semantic structures of the information and a crucial resource for high level clinical NLP applications, such as clinical text summarization [37] and knowledge base or knowledge graph construction and expansion [38–40]. Nevertheless, compared with the large number of packages for clinical concept extraction, there are only a few publicly available NLP systems have built-in modules for RE, such as cTAKES [41] and CLAMP [42]. To advance clinical RE, the clinical NLP community has organized several open challenges with shared tasks to solicit new state-of-the-art RE methods in the past decade. These challenges have focused on various medical relations, including drug-drug interaction (2013 SemEval) [9], treatment-test-problem relations (2010 i2b2) [12], and adverse drug events (2018 MADE1.0 and 2018 n2c2) [10,11]. These community efforts greatly improved the developments of RE benchmarks (i.e., publicly available clinical RE corpora) and methods in the medical domain.

In early RE studies, both rule-based and machine learning-based methods have been explored. For rule-based RE, dependency parsing [43], concept co-occurrence detection [44], and pattern matching [45] have been widely used. For machine learning-based approaches, researchers have explored support vector machines (SVMs), random forests (RF), and gradient boosting trees (GBT) for clinical RE [8]. For example, in the 2010 i2b2 challenge clinical RE task, Roberts et al. adopted the SVMs and achieved the best performance among all the challenge participants [46]; in the 2018 MADE1.0 challenge, Chapman et al. built an RF-based system and achieved the state-of-the-art performance [47]; and in the 2018 n2c2 challenge, Yang et al. developed a GBT-based approach, which achieved performance comparable to the best result [48]. Despite the excellent performance, most traditional machine learning-based RE methods still heavily rely on the lexical, syntactic, and semantic features manually crafted by domain experts, which is not only time-consuming but often not generalizable to new corpora.

Later, deep learning-based NLP models demonstrated the efficiency in learn linguistic features for downstream various tasks without human intervention, thanks to the emerging of word representation learning techniques (i.e., word2vec, fastText, GloVe) [49–52]. Deep learning models including both the long-short term memory architecture (LSTM; a special implementation of RNN) and convolution neural networks (CNNs) have been widely explored for RE. These deep learning-based systems demonstrated improved performances compared to traditional machine learning approaches on various benchmark datasets [18,19,53–55]. In the medical domain, different strategies have been proposed to integrate medical knowledge from diverse resources [56] to further improve clinical RE. For example, Bharath et al. designed a knowledge layer into an LSTM-based model to embed clinical knowledge features generated from the Unified Medical Language System (UMLS) [57] and demonstrated performance improvement. Another strategy widely used in the medical domain is to train word embeddings using clinical corpora such as the Medical Information Mart for Intensive Care III (MIMIC-III) notes [58], which has been adopted and achieved success for various clinical NLP tasks not limited to RE, including clinical NER, disease prediction, and computable phenotyping [59].

Most recently, various transformer-based NLP models (e.g., BERT, XLNet) have demonstrated superior performance for many NLP tasks. The key improvement of these transformers is the decoupling of language representation learning and specific NLP tasks training (e.g., classification) into two independent processes, i.e., pretraining and then fine-tuning. In the pretraining process, the transformer-based models are usually trained with a large amount of unlabeled text and optimized by language-modeling (LM) objectives, such as masked LM [32] and permutation LM [34]. In the fine-tuning process, these pretrained transformer models are then optimized towards specific NLP tasks (e.g., information extraction or question-answering) using annotated datasets. BERT is one of the pioneering NLP models that successfully implemented a transformer from prototype into a powerful language model. Pretrained with a corpus (combining the BooksCorpus [60] and English Wikipedia) with >3 billion words, BERT was fine-tuned to achieve state-of-the-art performances on many NLP benchmarking tasks in the general English domain [32]. Inspired by BERT, many transformer-based models with different architectures and/or pretraining strategies have been developed, such as RoBERTa and XLNet. These transformer models achieved state-of-the-art performance on many NLP benchmarks and have been applied in various NLP applications.

Nonetheless, most NLP research in the medical domain have heavily focused on the BERT model and there is no study that systematically examined other transformer models. Several studies examined the BERT LMs pretrained with clinical or biomedical corpus for different clinical NLP tasks, including concept extraction, question-answering, and text classification [25,26,29,30]. Not until recently, we systematically examined four transformer-based models for clinical concept extraction and demonstrate that RoBERTa-based models outperformed the BERT-based models on three clinical concept extraction benchmark datasets [15]. For clinical RE, there has been some research exploring BERT and its variants [61–65], but no study has systematically examined other transformers. One of the potential reasons is that there is a lack of publicly available software packages to facilitate clinical RE using various transformers. In this study, we systematically examined three widely used transformer-based NLP models (including BERT, RoBERTa, and XLNet), evaluated them using two benchmark datasets, and developed an open-source package with pretrained clinical transformers to facilitate the use of these state-of-the-art transformers for clinical RE.

## MATERIALS AND METHODS

### Dataset

In this study, we used two publicly available clinical RE datasets from the 2018 MADE1.0 challenge [10] and 2018 n2c2 challenge [11]. *Table 1* showed the descriptive statistics of the two datasets.

**Table 1.** Summary statistics of the clinical notes and annotated relations in the two datasets, where invalid relations were removed.

| Challenge dataset | Subset | Number of notes | Number of clinical relations |
|---|---|---|---|
| 2018 MADE1.0 | Training | 876 | 23,047 |
|  | Test | 213 | 4,128 |
| 2018 n2c2 | Training | 303 | 35,606 |
|  | Test | 202 | 23,462 |

The 2018 MADE1.0 dataset was curated from a collection of 21 cancer patients' clinical notes from the University of Massachusetts Memorial Medical Center, which was divided into a training set of 876 notes and a test set of 213 notes. There are seven types of relations in this corpus, including four relations for medication and medication attributes (Dosage, Route, Frequency, and Duration), two relations for medication and sign or symptoms (Adverse Drug Event [ADE] and Indication), and one relation for Severity. The 2018 n2c2 corpus has 505 discharge summary notes, which is a subset of the MIMIC-III; and the organizers released 303 notes as the training set and reserved the 202 notes for evaluation. Similar to the 2018 MADE1.0 dataset, the 2018 n2c2 corpus has eight relation categories. The Drug-ADE relation is for drugs and their associated adverse events and the Drug-Reason relation is for drugs and indications. The rest of the 6 categories are used for drugs and drug attributes (Strength, Duration, Route, Form, Dosage, and Frequency). In both datasets, the relations were annotated between two clinical concepts, which may be located in the same sentence or in different sentences. In the 2018 MADE1.0 training dataset, 80% of the relations were annotated in the same sentence and 20% of the relations were annotated across different sentences. In the 2018 n2c2 training dataset, the percentages became 90.8% and 9.2% for relations in the same sentence and different sentences, respectively.

**Relation extraction method**

The goal of RE is to identify concept pairs that have a valid relation and determine the category of the relation. This study focuses on clinical RE, therefore, we did not include the clinical concept extraction models, which have been systematically examined in our recent study [15]. In the following experiments, we assume that all clinical concepts have been identified from the clinical text. We systematically examined several key aspects of clinical RE, including methods to generate candidate concept pairs, 3 transformer-based NLP models, 2 classification strategies (binary *vs* multi-class classification), and different strategies to handle cross-sentence relations.

*Candidate Concept Pair Generation*

A critical step of RE is to identify the concept pairs with a relation, which is typically formed on two concepts. A straightforward way of generating concept pairs is to enumerate all combinations between all concepts within a document. However, this method often brings too many negative samples and leads to an extremely imbalanced positive-negative sample ratio as the combination space is often large. In previous studies [48,66], we have explored heuristic rule-based strategies to reduce the generation of negative samples. In this study, we also applied a similar heuristic rule: two concepts can be a candidate pair only if there is a relation category defined between the semantic categories of the two concepts. For example, A Drug concept and a Dose concept can form a candidate pair, but an ADE concept and a Dose concept cannot possibly form a candidate pair as there is no valid relation can be defined between them.

*Transformer-based RE Models*

In this study, we systematically examined three transformer-based language models, including BERT [32], RoBERTa [33], and XLNet [34] for clinical RE. We utilized transformers for contextual representation learning and added a soft-max layer for the classification task.

BERT is a bidirectional transformer-based encoder language model pretrained over a large general English domain corpus. BERT adopted the masked language modeling (MLM) and next-sentence prediction (NSP) training objectives to create deep representations capturing contextual information. BERT has two variants featuring different model sizes, including a

BASE version (110 million parameters) with 12 transformer layers, 768 hidden units and 12 attention heads, and a LARGE version (340 million parameters) with 24 transformer layers, 1024 hidden units and 16 attention heads. RoBERTa is a transformer-based language model shared the same architecture as BERT but pretrained with modified strategies. Unlike BERT's static MLM (masking pattern generated during preprocessing), the RoBERTa pretraining adopted a dynamic MLM where masking patterns were generated during the training with different random seeds to avoid using the same masks for each training instance in every epoch. Other modifications included removing NSP during the pretraining and using a 10 times larger general domain English training corpus compared to BERT. XLNet is a generalized auto-regressive language model [67,68] pretrained with the permutation language modeling (PLM) objective. While MLM aims to reconstruct the original data from corrupted input, the PLM seeks to estimate the probability of a token conditioned on all permutations of the tokens in a sequence. Same as BERT, XLNet also has the BASE version and LARGE version with 110 million and 340 million parameters, respectively.

Previous studies [15,26,30] have demonstrated that transformer-based models pretrained with clinical corpora (e.g., the MIMIC-III clinical notes) could further improve the model performances for downstream NLP tasks in the medical domain. Thus, we compared general transformer models pretrained using general English corpora with clinical transformers pretrained using clinical text. For general transformer models, we examined both the BASE and LARGE variants (i.e., BERT-base and BERT-large). For clinical pretrained models, we adopted the clinical transformers developed in our previous study pretrained using the MIMIC-III corpus [15] (e.g., BERT-clinical).

*Training strategies*

For each transformer model, we compared two training strategies, including (1) a multi-class classification strategy, where we treat each relation category as a unique class and added a "*non-relation*" class when there is no relation between two concepts; then we developed a unified transformer model to classify the candidate pairs into one of the predefined relation categories, which is a traditional multi-class classification problem; and (2) a binary classification strategy, where we applied a 2-stage classification procedure consists of a binary classification to determine whether the candidate pair has a relation (positive) or not (negative) and then a rule-based pipeline to infer the relation category according to the semantic categories of the two concepts. For example, if a concept pair was classified as positive and one of the entities is Drug and the other is ADE, then the rule-based pipeline will infer that the relation between the two concepts is a Drug-ADE relation.

*Strategies to handle cross-sentence relations*

A critical challenge of RE is how to handle cross-sentence relations, i.e., relations formed between concepts located in two different sentences. To better categorize cross-sentence relations, we define the cross-sentence distance (CSD) as the number of sentence boundaries between the two entities in a relation. For example, the CSD is 0 if the two concepts are in the same sentence; and the CSD would be 1 if the two concepts occurred in consecutive sentences. In our previous study [48], we found that training multiple classifiers for each group of relations with the same CSD could further improve the performance of RE when using a traditional machine learning model, i.e., the gradient boosting tree (GBT). Therefore, we systematically compared two strategies to handle cross-sentence relations for transformers, including (1) a *UNIFIED* model – we train a unified transformer model to classify all relations regardless of they are cross-sentence or not; and (2) a *DISTANCE-SPECIFIC* model – for each group of concept pairs with the same CSD, we train an individual transformer model. According to our previous study, we systematically explored the relations with CSD from 0 – 4 as there is no performance improvement when considering candidate pairs with CSD > 4. Therefore, for each transformer-based RE model, we developed 4 classifiers to classify candidate concept pairs with CSD from 0-4, respectively.

*The architecture of the transformer-based clinical RE*

**Figure 1** shows an overview of the transformer-based RE model architecture (using BERT as an example in the figure), where $Tok_i$ is the *i*-th token of the input, $Emb_i$ is the embedding of $Tok_i$, and $T_i$ is the fine-tuned contextual representation of $Tok_i$ leaned by the transformer. The transformer-based models also introduce several special tokens to (1) learn representations of the entire input; and (2) format raw inputs (e.g., separate or padding sentences). For example, BERT inserts the [CLS] token at the beginning of the input to learn sentence-level representation and insert the [SEP] token at the middle and end of the input to define sentence boundaries. Most transformer-based models can take two sentences separated by the special separation token (e.g. [SEP] in Figure 1) as input to learn their joint representations (i.e., two-sentence input scheme), which can be used to model the relation classification - the two sentences containing the two concepts were used as the input to learn a joint representation, which was then used by the classification layer for relation classification. Therefore, the RE model can learn both individual contextual information of each entity and their joint representation for relation type classification. To distinguish relation entities from other tokens in sentences, we introduced two sets of entity markers (i.e., [S1], [E1] and [S2], [E2]) to define entity spans. If the two concepts in a relation locate in the same sentence, then the two input sentences are the same but with different markers

(e.g., [S1] and [E1] in the first sentence; [S2] and [E2] in the second sentence). To determine the relation type, we concatenated the contextual representations of the [CLS] token and all four entities markers (i.e., $T_{cls}$, $T_{S1}$, $T_{E1}$, $T_{S2}$, and $T_{E2}$), and fed this representation to a classification layer to calculate the probability for each relation category.

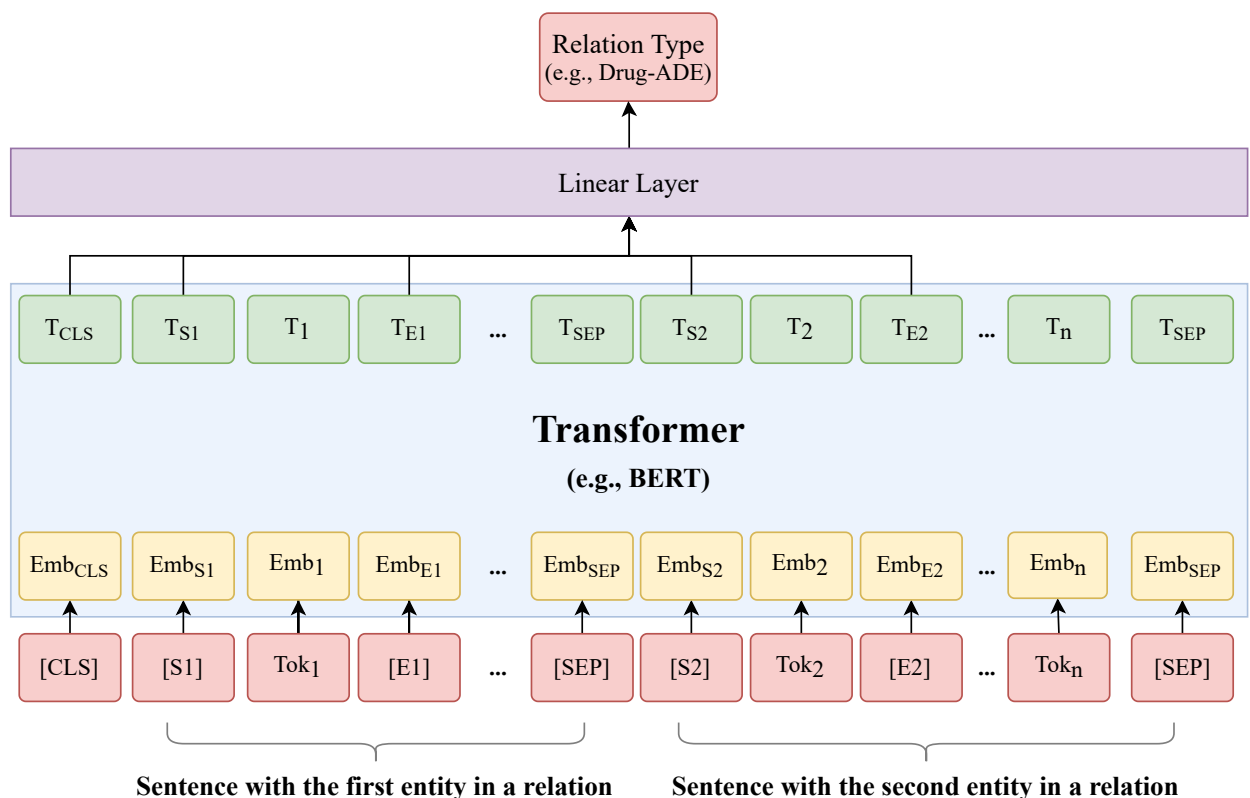

**Figure 1.** The overview of the relation identification model and workflow (using BERT as example).

Transformers typically learn representations at various levels (e.g., token-level and sentence-level). How to use these representations for relation classification is still unclear. In this study, we compared four representation schemes including (1) only using the classification token representation (i.e., [CLS] in BERT), (2) using representations of the classification token and entity start markers (i.e., [S1], [S2]), (3) using representations of the classification tokens and all entity markers (including start markers and end markers as shown in Figure 1), and (4) only using representations of entity start markers.

**Experiments and evaluation**

*Experiment set up:* We developed the RE models on top of the existing transformer architectures implemented in the Transformers library developed by the HuggingFace team [69] using PyTorch [70]. We used the general transformer models pretrained using general English text from the HuggingFace model repository. For the clinical transformers, we adopted the models we pretrained using the MIMIC III corpus (BERT-clinical, RoBERTa-clinical, and XLNet-clinical) in our previous study [15]. For fine-tuning, we adopted a five-fold cross-validation strategy to optimize the model hyperparameters, including the training epoch number (in a range from 3 to 6) and training batch size (4, 8, and 16). We kept all other hyperparameters as default (e.g., learning rate at 1e-5 and random seed at 13) during the experiments. The best models were selected according to the cross-validation performances measured as micro-averaged strict F1-score. All experiments were conducted using five Nvidia A100 GPUs.

*Evaluation metrics*: Following the standard evaluation of clinical RE, we compared the performance of transformer-based models using the strict micro-averaged precision, recall, and F1-score aggregated from all relation categories. We used the official evaluation scripts provided by the 2018 n2c2 challenges to calculate these scores. The entities involved in the evaluation were from the gold standard annotations.

**RESULTS**

Table 2 shows the performance of all transformer-based models for the 2018 MADE1.0 and 2018 n2c2 datasets under the UNIFIED setting. For the 2018 MADE1.0 dataset, the RoBERTa-clinical achieved the best F1-score of 0.8959 using the binary classification strategy and the XLNet-clinical achieved the best F1-score of 0.8919 for the multi-class classification strategy.

Both of the two models achieved a significantly higher performance compared to the best F1-score of 0.8684 reported in the 2018 MADE1.0 challenge [10]. For the 2018 n2c2 dataset, both RoBERTa-clinical and XLNet-clinical achieved the best F1-score of 0.9606 using binary classification strategy, which is comparable to the best performance reported in the 2018 n2c2 challenge (0.9630). For binary classification and multi-class classification strategies, we observed that almost all transformer models using binary classification strategy outperformed the corresponding models using multi-class classification strategy. On average, the binary classification strategy outperformed the multi-class classification strategy by ~0.3% and ~1.3% on the 2018 MADE1.0 and 2018 n2c2 datasets, respectively. All transformers achieved the best results by only considering candidate pairs with CSD of 0 and 1. The performance dropped greatly when we add the candidate pairs with CSD > 1.

**Table 2.** The performance of RE on the 2018 MADE1.0 and 2018 n2c2 datasets using UNIFIED setting.

| Corpus | Model | Evaluation Metrics | | | | | |
| --- | --- | --- | --- | --- | --- | --- | --- |
| | | Binary Classification | | | Multi-class Classification | | |
| | | Precision | Recall | F1 | Precision | Recall | F1 |
| **2018 MADE1.0** | | | | | | | |
| | BERT-base | 0.8938 | 0.8669 | 0.8801 | 0.8997 | 0.8487 | 0.8734 |
| | BERT-large | 0.9047 | 0.8648 | 0.8843 | 0.9112 | 0.8576 | 0.8836 |
| | BERT-clinical | 0.9022 | 0.8688 | 0.8852 | 0.9010 | 0.8616 | 0.8809 |
| | RoBERTa-base | 0.8986 | 0.8705 | 0.8843 | 0.9015 | 0.8597 | 0.8801 |
| | RoBERTa-large | **0.9131** | 0.8710 | 0.8916 | **0.9214** | 0.8532 | 0.8860 |
| | RoBERTa-clinical | 0.9126 | **0.8799** | **0.8959** | 0.9092 | **0.8727** | 0.8906 |
| | XLNet-base | 0.9067 | 0.8640 | 0.8849 | 0.9030 | 0.8518 | 0.8766 |
| | XLNet-large | 0.9081 | 0.8638 | 0.8854 | 0.9153 | 0.8669 | 0.8905 |
| | XLNet-clinical | 0.9070 | 0.8736 | 0.8900 | 0.9149 | 0.8700 | **0.8919** |
| **2018 n2c2** | | | | | | | |
| | BERT-base | 0.9589 | 0.9472 | 0.9530 | 0.9463 | 0.9336 | 0.9399 |
| | BERT-large | 0.9642 | 0.9481 | 0.9561 | 0.9484 | 0.9305 | 0.9394 |
| | BERT-clinical | 0.9597 | 0.9540 | 0.9568 | 0.9527 | 0.9401 | 0.9463 |

|  | RoBERTa-base | 0.9615 | 0.9483 | 0.9549 | 0.9537 | 0.9332 | 0.9434 |
|  | RoBERTa-large | 0.9629 | 0.9512 | 0.9571 | 0.9551 | 0.9369 | 0.9459 |
|  | RoBERTa-clinical | 0.9701 | 0.9512 | **0.9606** | 0.9565 | 0.9376 | **0.9470** |
|  | XLNet-base | 0.9643 | 0.9486 | 0.9563 | 0.9519 | 0.9344 | 0.9431 |
|  | XLNet-large | **0.9711** | 0.9474 | 0.9591 | **0.9569** | 0.9357 | 0.9462 |
|  | XLNet-clinical | 0.9686 | **0.9526** | **0.9606** | 0.9524 | **0.9408** | 0.9466 |

\* Best scores are highlighted as bold for each classification method in each dataset. All models achieved the best results by considering candidate relations with CDS of 0 and 1.

Table 3 summarized the performances (F1-scores only; see supplement material for precision and recall) for all transformer-based RE models under the DISTANCE-SPECIFIC setting. For each transformer, we trained 10 models (5 for each group of candidate pairs with the same CSD from 0-4 combined with 2 classification strategies [binary and multi-class]). Here, we only reported the best F1 score and associated CSD (e.g., 0-2 denote the best F1-score was achieved by considering candidate pairs with CSD of 0, 1, and 2). On the 2018 MADE1.0 test dataset, the RoBERTa-clinical model achieved the best performances for both multi-class and binary classification strategies (0.8898 and 0.8924) by considering CSD from 0-4. For the 2018 n2c2 test dataset, the XLNet-clinical model achieved the best F1-score of 0.9486 and 0.9610 for the multi-class and binary strategies by considering candidate pairs with CSD from 0-2. On average, the binary classification strategy outperformed the multi-class classification strategy by ~0.005 and ~0.015 on the two datasets, respectively.

**Table 3.** The performance of RE on the 2018 MADE1.0 and 2018 n2c2 datasets using distance-SPECIFIC setting (only F1-scores).

| Model | 2018 MADE1.0 | | 2018n2c2 | |
|---|---|---|---|---|
|  | **Multi-class** | **Binary** | **Multi-class** | **Binary** |
| BERT-base | 0.8793 (0-4) | 0.8824 (0-4) | 0.9361 (0-2) | 0.9537 (0-3) |
| BERT-large | 0.8863 (0-4) | 0.8800 (0-4) | 0.9431 (0-2) | 0.9561 (0-2) |
| BERT-clinical | 0.8862 (0-4) | 0.8841 (0-4) | 0.9473 (0-2) | 0.9580 (0-2) |
| RoBERTa-base | 0.8786 (0-4) | 0.8825 (0-4) | 0.9410 (0-2) | 0.9536 (0-2) |
| RoBERTa-large | **0.8898** (0-4) | 0.8855 (0-3) | 0.9444 (0-1) | 0.9550 (0-1) |
| RoBERTa-clinical | **0.8898** (0-4) | **0.8924** (0-4) | 0.9483 (0-2) | 0.9592 (0-2) |
| XLNet-base | 0.8794 (0-4) | 0.8876 (0-4) | 0.9421 (0-2) | 0.9561 (0-2) |
| XLNet-large | 0.8843 (0-3) | 0.8853 (0-3) | 0.9447 (0-1) | 0.9584 (0-2) |
| XLNet-clinical | 0.8837 (0-3) | 0.8916 (0-3) | **0.9486** (0-2) | **0.9610** (0-2) |

\* (1) All reported scores are best F1-scores; (2) see supplement for detailed precisions, recalls; (3) cross-distances for the best F1 score are reported as ranges in the parentheses; and (4) the best F1 scores were highlighted as bold.

Table 4 compared different strategies to combine the representations learned by the transformers using the top-performed transformer, RoBERTa-clinical, with binary classification strategy and the UNIFIED setting. On both datasets, scheme-3 (i.e., representations of classification token and all entity markers) achieved the best performances. We observed that the performance improvement by scheme-3 was mainly from the improvement in precision. Scheme-2 and 4 achieved similar performance on both datasets and scheme-1 (i.e., only using classification token representation) is the lowest (0.8824 and 0.9585) for both datasets.

**Table 4.** Results for different representation schemes achieved by the RoBERTa-clinical with the binary classification strategy and the UNIFIED setting.

| **Relation Representation Scheme** | **2018 MADE1.0** | | | **2018 n2c2** | | |
|---|---|---|---|---|---|---|
| | Precision | Recall | F1 | Precision | Recall | F1 |
| **1** classification token only | 0.8896 | 0.8753 | 0.8824 | 0.9644 | **0.9526** | 0.9585 |
| **2** classification token and entity start markers | 0.9098 | 0.8724 | 0.8907 | 0.9665 | 0.9518 | 0.9591 |
| **3** classification token and all entity markers | **0.9126** | **0.8799** | **0.8959** | **0.9701** | 0.9512 | **0.9606** |
| **4** only entity start markers | 0.9054 | 0.8734 | 0.8891 | 0.9675 | 0.9513 | 0.9593 |

\* Best precision, recall, and F1 are highlighted as bold.

## DISCUSSION AND CONCLUSION

Clinical RE is a challenging task to identify medical relations from clinical text, which is important for constructing comprehensive patient profiles from EHRs. In this study, we explored three widely used transformer-based models (i.e., BERT, RoBERTa, and XLNet) for clinical RE using two benchmark datasets (i.e., 2018 MADE1.0 and 2018 n2c2). We systematically compared several critical aspects of clinical RE including 2 classification strategies (i.e., binary vs. multi-class), 2 training settings to handle cross-sentence relations (i.e., UNIFIED vs. DISTANCE-SPECIFIC), and 4 schemes to combine the representations learned by transformers for relation classification. This study demonstrated the efficiency and discovered many critical details in using transformer-based models for clinical RE, which will greatly help others design transformer-based solutions for their needs.

Two transformers achieved state-of-the-art performance, including RoBERTa-clinical, which achieved the best performance of 0.8959 (compared to the best F1-score of 0.8684 in the challenge) for the 2018 MADE1.0 dataset, and XLNet-clinical, which achieved the best performance of 0.9610 for the 2018 n2c2 dataset. These experimental results showed that transformers pretrained using clinical text outperformed those pretrained using general English text for RE, which is consistent with findings for the clinical concept extraction task [15]. On average, the binary classification strategy is better than the multi-class classification strategy. For example, both RoBERTa-clinical and XLNet-clinical achieved the best performance for the two datasets by using the binary classification strategy. One potential reason is that the binary strategy combined all candidate pairs into a much bigger group (i.e., the positive group) that improved the sample size in the classification task compared with the multi-class strategy. Among the three transformers, the RoBERTa-based and XLNet-based models generally achieved good and robust performance for RE. The BERT-based models, which achieved many state-of-the-art results for clinical concept extraction, may not be the best solution for clinical RE.

For the two strategies to handle cross-sentence relations, our results suggest that there is no significant difference between the UNIFIED and DISTANCE-SPECIFIC settings in terms of the overall F1-score. For example, on the 2018 n2c2 dataset, the best F1-score using UNIFIED was 0.9606 and that using DISTANCE-SPECIFIC was 0.9610. However, there are significant differences in handling individual cross-sentence relations. For example, all transformer models achieved the best performance by considering the candidate pairs with CSD <2 under the UNIFIED setting. There is no further improvement when adding candidate pairs with CSD >1. In fact, some of the models' performance actually dropped as more negative samples were brought into the training process. On the other side, the DISTANCE-SPECIFIC setting could better handle cross-sentence relations with CSD >1. For example, when using the DISTANCE-SPECIFIC setting, many transformers could still get better performance by considering candidate pairs with CSD from 0-4; however, this performance improvement did not contribute to the overall F1-score as there were more negative samples brought into the training. For example, when generating candidate pairs with CSD=2 in the 2018 n2c2 training dataset, there are 54,012 negative samples generated along with only 446 positive samples. Generally speaking, the candidate pairs with CSD from 0-1 are the most important as most of the relations were formed in the same or consecutive sentences. For example, if we only include candidate relations with cross-sentence distance = 0, the XLNet-clinical model only achieved an F1-score of 0.9381 (compared with 0.9610, more details in the supplement material). Our results suggest that the DISTANCE-SPECIFIC setting can handle more cross-sentence relations but brings more noise at the same time, indicating it is still challenging to handle cross-sentence relations in clinical RE.

This study examined how to effectively use the representations generated by transformers for clinical RE. A recent study [71] screened several methods to generate relation representations from the BERT encoder and reported that the method using the representations of the entity start markers as the relation representations achieved significantly better performance compared to other methods. However, in the biomedical domain, most transformer-based RE studies only used the representation of the special classification token (e.g., [CLS] for BERT) to determine the relation types [25,29,61,72]. In this study, we systematically compared 4 different representation schemes (Table 4) and found that scheme 3 consistently achieved the best performance for clinical RE. Our study shows that the positional information (i.e., entity markers) is critical for transformers to learn useful representations for RE. Although the classification token is originally designed to capture the sentence-level representation, this study shows that the sentence-level representation can help RE as well.

## LIMITATIONS

This study has limitations. The binary classification strategy explored in this study is not generalizable to complex settings where there are multiple types of relationship defined between two concepts. We did not explore methods that can alleviate the imbalanced negative-positive sample issue when adding candidate pairs with CSD >1. Future studies are needed to better handle the cross-sentence relations and avoid generating too many negative samples.

## ACKNOWLEDGMENTS

We would like to thank the 2018 MADE1.0 and 2018 n2c2 challenge organizers to provide the annotated corpora. We gratefully acknowledge the support of NVIDIA Corporation and NVIDIA AI Technology Center with the donation of the GPUs and the computing resources used for this research.

## FUNDING STATEMENT

This study was partially supported by a Patient-Centered Outcomes Research Institute® (PCORI®) Award (ME-2018C3-14754), a grant from National Institute on Aging 1R56AG 069880, a grant from the National Cancer Institute, 1R01CA246418 R01, a grant from CDC (Centers for Disease Control and Prevention) 1U18DP006512-01, the University of Florida (UF)

SEED Program (DRPD-ROF2020, P0175580), and the Cancer Informatics and eHealth core jointly supported by the UF Health Cancer Center and the UF Clinical and Translational Science Institute. The content is solely the responsibility of the authors and does not necessarily represent the official views of the funding institutions.

## COMPETING INTERESTS STATEMENT

Xi Yang, Zehao Yu, Yi Guo, Jiang Bian, and Yonghui Wu have no conflicts of interest that are directly relevant to the content of this study.

## CONTRIBUTORSHIP STATEMENT

XY, JB and YW were responsible for the overall design, development, and evaluation of this study. ZY was involved in the results analysis. XY, YG, JB and YW did the initial drafts and revisions of the manuscript. All authors reviewed the manuscript critically for scientific content, and all authors gave final approval of the manuscript for publication.

## SUPPLEMENTARY MATERIAL

Supplementary material is available at *Journal of the American Medical Informatics Association* online.

The software package is publicly available under the MIT license at https://github.com/uf-hobi-informatics-lab/ClinicalTransformerRelationExtraction.

8/16/21 11:56:00 AM